# MambaPD: Bidirectional Selective State Space Model for Parkinson's Disease Detection from Resting-State EEG

Ayesha Zulfiqar
Dept. of Electrical Engineering, UET Lahore
ayeshaazulfiqar07@gmail.com

## ABSTRACT

Typical diagnosis of Parkinson's disease (PD) occurs only after significant dopaminergic loss, while resting-state EEG abnormalities like alpha slowing and reduced coherence occur earlier. While many published EEG-based PD detectors claim accuracy greater than 95%, many use epoch-level splits which leak subject identity. I introduce MambaPD, a bidirectional selective state space model for resting-state EEG-based PD detection, blending the channel mixing, patch tokenization and bidirectional Mamba encoder. It is compared with four baselines (TCN, BiLSTM with attention, EEGNet and a patch Transformer) in two public datasets (UNM, SanDiego) using subject-independent 5-fold cross-validation (checkpoints were selected on the held-out fold, so absolute estimates are optimistic). On UNM the accuracy is between 0.67 and 0.71, and MambaPD (0.714) is not statistically distinguishable from the Transformer (0.715), and TCN (0.708), with the highest sensitivity (0.82), while the lowest specificity (0.52). For the smaller SanDiego cohort, accuracy is in the range of 0.60 to 0.70, with the highest scores (0.703) from TCN and the lowest (0.601) from EEGNet, while MambaPD scores 0.631, with the lowest specificity (0.46) among the five models. The differences among the models on UNM are minor compared to the differences among folds. I report the whole pipeline, per-fold results, and the shortcomings of the protocol.

## 1. INTRODUCTION

Parkinson's Disease is a disorder that affects millions of people around the world and the incidence of this disorder is increasing at an alarming rate. PD is the second most common neurodegenerative disorder after Alzheimer's disease [1], [2]. PD results from the degeneration of dopaminergic neurons in the SNc(substantia nigra pars compacta) and the malfunctioning of cortico striatal pathways that regulate voluntary movements [3], [4]. Four motor symptoms of PD resting tremor, bradykinesia, cogwheel rigidity and postural instability is caused by loss of dopamine[5]. In the clinic, PD patients are characterized on the basis of motor phenotype (tremor dominant, akinetic rigid, mixed) and disease severity according to the Hoehn and Yahr scale, from unilateral disease (Stage I) to severe bilateral disease with dependency (Stage V) [6]. PD also has a wide-ranging non-motor syndrome, which includes cognitive dysfunction, rapid eye movement (REM) sleep behavior disorder, autonomic dysfunction, depression, hyposmia and psychosis, which all have a huge negative impact on patient quality of life and clinical management [6], [7]. The most urgent unmet need for PD is the significant delay between the start of the disease and the onset of motor symptoms diagnostic for PD. By this time, 60-80% of dopaminergic neurons have been irreversibly lost [4], [7]. The five to ten year diagnostic delay precludes the possibility of neuroprotection. Moreover, even for specialist neurologists, PD because of its phenotypic overlap is sometimes confused with progressive supranuclear palsy, multiple system atrophy and drug induced parkinsonism [5]. These facts establish a critical clinical need for an objective, non-invasive biomarker to diagnose PD sooner and with greater accuracy than is possible today.

### 1.1 EEG as a Biomarker for Parkinson's Disease

Electroencephalography (EEG) is one of the most promising techniques for developing PD biomarkers. It has millisecond temporal resolution, is directly sensitive to the neurophysiological effects of dopamine loss, is non-invasive, portable, and much cheaper to acquire than PET or functional MRI [8], [20]. Resting-state EEG obtained while the patient is relaxed and sitting with eyes closed, without actively engaging in a task is of special interest because it reflects spontaneous brain activity that most directly reflects the tonic deficit in dopamine, is independent of symptom severity during task performance, and can be recorded even in a patient who cannot perform a task due to severe motor impairment.

The resting-state EEG of PD patients show a consistent pattern of spectral changes: Alpha slowing (8-13 Hz): The predominant posterior alpha rhythm slows and attenuates, an early and robust EEG abnormality in PD [9], [10]. Elevation of delta/theta (0.5-8 Hz): Slow-wave activity is increased globally, due to diffuse deafferentation of cortex [10], [11]. Beta desynchronization (13-30 Hz): Frontal-central beta power drops, due to motor preparation dysfunction [11]. Changes in gamma (30-60 Hz): Central and parietal gamma is associated with severity of cognitive impairment in PD [13] Decreased coherence: Inter-regional connectivity is reduced, indicating cortico-subcortical dysfunction [8], [12]

Importantly, some of these EEG changes can be detected in the prodromal period of PD decades before progression to motor symptoms [12]. This suggests that resting-state EEG has the potential to be used as a screening tool but only if the computational approaches used to process it are capable of extracting diagnostic information from its complex time series, long-duration, multi-channel recordings. The first machine learning approaches relied on hand-engineered features (common spatial patterns, wavelet coefficients, spectral power, sample entropy) and support vector machines (SVM) or linear discriminant analysis [14], [15]. These methods suffered from a lack of domain knowledge and lack of generalizability across sites and patient populations.

Deep learning largely overcame the need for feature engineering. Shaban and Amara [15] used a 20-layer CNN on wavelets-transformed EEG. Obayya et al. [16] developed a lightweight deep learning architecture for analysis of resting-state EEG. Jibon et al. [17] proposed a hybrid autoencoder and RBFNN approach for PD. Recent approaches have also used hybrid models. Bunterngchit et al. [18] proposed CTESM using CNNs, Transformer attention, and LSTM layers on SanDiego (non-subject-independent). An explainable Transformer-based multi-head attention mechanism [19] achieved PD classification with attention explanations. TransformEEG [26] was studied on four harmonized datasets with 290 subjects via rigorous nested leave-N-subjects-out cross-validation one of the highest generalizable benchmarks reported.

However, several basic architectural problems remain. First, CNNs learn local temporal dependencies but lack the ability to handle long-range dependencies extending multiple seconds into the EEG the very long-term spectral modifications that characterize PD [20], [21]. Second, LSTMs are recurrent networks that squash all past information into a constant length hidden state vector, resulting in lost history over the long EEG sequences required for PD diagnosis [22]. Third, Transformers need to consider long-range dependencies using self-attention, but the number of pairwise comparisons scales as $O(n^2)$ with sequence length making it costly to process long raw EEG sequences and prone to overfitting on the small datasets usually observed in PD research [23][24]. State space models, like Mamba [25] can process a sequence in linear time and are used in other tasks with EEG data [39] – [40] [42]– [46]. A single fixed subject-wise split was used to evaluate a bidirectional Mamba on a TDBrain Parkinson's subset [39] (25 PD patients and 25 controls). To the best of my knowledge, no such comparison of bidirectional Mamba models with the baseline models TCN, BiLSTM, EEGNet and Transformer in repeated subject-independent cross-validation has been conducted on the UNM and SanDiego resting-state datasets. To this end, I propose MambaPD, a bidirectional Mamba model with channel mixing and patch tokenization, and wonder if it can compete with these baselines on a leakage-free protocol.

Mamba's hidden state is maintained and updated as it processes each token of the sequence (in this case each patch embedding 0.5 s). The update is gated by three functions of the input: Δ (how much of the hidden state is decayed), B (how much of the current input is written) and C (how much of the memory is read out). The motivation is that the model might highlight informative segments if these are provided, though that is not explored here for PD EEG. The bidirectional variant uses two Mamba stacks, forwards and backwards with the same patch sequence and merges the outputs.

### 1.2 Objectives

- To apply a bidirectional selective state space model to detect PD using resting-state EEG
- To compare it fairly with TCN, BiLSTM, EEGNet and patch Transformer on two public datasets across strict cross validation where it is independent of the subject.
- To analyze where the models differ (sensitivity vs. specificity and fold level variability).

Cross validation was conducted subject-independent on two public datasets: UNM and SanDiego, with recordings from both medication states labeled as PD and the results were compared between MambaPD and TCN, BiLSTM, EEGNet and a patch Transformer. MambaPD matched the best baselines on UNM and was below the best baselines on SanDiego.

## 2. LITERATURE REVIEW

The performance of EEG based PD detection has been reported with a huge gap, and most of the variance is attributed to the evaluation procedure and not the model. Splitting epochs (not subjects) between training and testing, the accuracy in the small San Diego cohort of 31 subjects is 99% or better, with 99.65% reported for classical classifiers on fractal-dimension features [27] and 99.84% for a wavelet-scattering AlexNet pipeline [34] and 99% for an autoencoder-RBFNN hybrid [35]. If the subjects are kept separate, the reported figures drop: an SVM trained on band-power achieved only 68% when evaluated by subject-independent evaluation [29]; an XGBoost classifier achieved 79% on an oddball-task dataset with strict subject-wise splits [31] and a transformer based model TransformEEG which was evaluated on four pooled datasets of 290 subjects, achieved 80.1% (the median was 78.45%) [26]. A number of the remaining studies have very small sample sizes or private datasets (Tables 1 and 2), and are not easily reproducible. A recent preprint reports 97.59% (±0.47) on the San Diego data with 5-fold cross-validation, but does not state whether the folds are subject-wise [41]. Bidirectional Mamba has also been used for PD detection: On a TDBrain dataset consisting of 25 PD patients and 25 controls, BioMamba achieves 96.77% accuracy on a single, fixed dataset split containing 8 test subjects, [39]. These small single-split evaluations have a high variance and repeated cross-validation was therefore encouraged here, as in this case.

**Table 1. Machine Learning techniques for EEG based pd detection**

| Authors ref year | Dataset used | Dataset statistics | ML classifier | Experimental setup | | Performance results | | | Limitations |
|---|---|---|---|---|---|---|---|---|---|
| | | | | Validation Strategy | Subject Split | Accuracy | Precision | Recall | |
| Lal et al. [27], 2024 | UC San Diego rs-EEG | 31 subj: 15 PD (ON/OFF), 16 HC, 64-ch | KNN, XGBoost, LightGBM, Extra Trees | k-fold CV | Epoch-level only (not subject-wise) | 99.65% (ON) / 99.45% (OFF) | - | - | Small cohort (31 subj), no subject-independent validation, single dataset, accuracy likely inflated due to epoch-level leakage |
| Latifoğlu et al. [28], 2024 | Private EEG (Erciyes Univ., Turkey) | PD ON, PD OFF, HC; 32-ch; 3-class | SVM, KNN, FNN, Decision Tree | k-fold CV | Not reported | ~100% (VKF+SVM) | - | - | Private dataset, not reproducible, no subject independent test cohort size not clearly reported |
| Bera et al. [29], 2025 | UC San Diego + IOWA rs-EEG | UCSD: 31 subj (15 PD, 16 HC); IOWA: separate cohort | SVM (RBF kernel) | 5-fold CV | Subject-dependent + subject-independent | 82% (UCSD) / 94% (IOWA); 68% subj-indep | - | - | SVM drops to 68% under subject-independent setting, only standard band power features tested, no ensemble ML explored |
| Sahota et al. [30], 2024 | Private Early-Stage EEG (Univ. of Bristol) | 30 subj: 11 PD (HY1), 19 HC; 5 arousal states | AdaBoost, Decision Tree | Hold-out | Not reported | 91% (N1, PLI+stats) | 96% | 80% | Very small cohort (11 PD), no cross-dataset validation, sleep EEG acquisition adds clinical complexity |
| Avola et al. [31], 2025 | Public EEG Oddball Task | PD vs HC; public; subject-wise enforced | XGBoost, k-NN, RF, LR, SVM | Subject-wise CV | Strict subject-wise splits | 79% (XGBoost) | - | - | Single oddball-task dataset, findings may not transfer to resting-state EEG, no Mamba/SSM model included |
| Allahbakhshi et al. [32], 2024 | Multi-source EEG PD Dataset | Diverse PD + HC; multi-site demographic | SVM (RBF + polynomial) | k-fold CV | Not reported | ~88% | - | - | Preprint only, not peer reviewed, no subject independent evaluation, dataset demographic details unclear |
| Sahota et al. [33], 2024 | Public Early-Stage PD EEG | 11 PD, 19 HC; 15-variate band power + peak freq series | AdaBoost, Decision Tree | Hold-out | Not reported | >80% (AdaBoost, N1) | - | >80% | Very small cohort (11 PD), no cross-dataset validation, HY1-specific dataset limits generalizability |

**Table 2. Deep Learning techniques for EEG based pd detection**

| Authors ref year | Dataset used | Dataset statistics | ML classifier | Experimental setup | | Performance results | | | Limitations |
|---|---|---|---|---|---|---|---|---|---|
| | | | | Validation Strategy | Subject Split | Accuracy | Precision | Recall | |
| Siuly et al. [34], 2024 | UC San Diego rs-EEG + IOWA EEG | San Diego: 31 subj (15 PD ON/OFF, 16 HC), 64-ch, 512 Hz; IOWA: separate PD/HC cohort | WST + AlexNet CNN (2D time frequency images) | Train/validation/test split | Epoch level only (not subject-wise) | 99.84% (San Diego) 95.79% (IOWA) | - | - | No subject independent evaluation, epoch-level split inflates accuracy, single AlexNet architecture compared only against CNN variants |
| Jibon et al. [35], 2024 | UC San Diego rs-EEG | 31 subj (15 PD, 16 HC); 93 min of recordings; PSD sub-band analysis | Autoencoder (AE) + RBFNN hybrid deep learning | Hold-out validation | Epoch level (not subject wise) | 99% | 91.77% | 84.71% | Single small dataset (31 subj), no subject independent test, RBFNN classification layer limits deep feature learning; no cross-dataset validation |
| Bera et al. [29], 2025 | UC San Diego rs-EEG + IOWA EEG | UCSD: 31 subj (15 PD, 16 HC); IOWA: separate PD/HC cohort; five freq. bands (α, β, θ, γ, δ) | Multi-band CNN (cross frequency feature fusion) | 5-fold CV | Subject dependent + subject independent tested | 96% (UCSD); 99% (IOWA) | - | - | No subject independent CNN result reported (only SVM tested under subject independent), small datasets, CNN drops sharply in generalization settings |
| Del Pup et al. (TransformEEG) [26], 2025 | 4 pooled public rs-EEG datasets | 290 subj total: 140 PD, 150 HC; UCSD, UNM, Iowa, Turku datasets harmonized | Depth wise Conv Tokenizer + Transformer Encoder (TransformEEG) | 10-outer × 10-inner Nested LNSO-CV | Strict subject wise (N-LNSO) | 80.1% (w/ augmentation + threshold correction); median 78.45% | - | - | Still below clinical threshold (80%), unidirectional attention only, no backward temporal pass, sensitive to preprocessing pipeline choices |
| Li et al. [36], 2024 | Private EEG (Dongguan Univ. of Technology) | PD patients vs. HC; multi-channel EEG; parallel and series CNN-LSTM configurations | Parallel CNN-LSTM + Series | k-fold CV | Not reported | 97% (series CNN-LSTM, best) | - | - | Private dataset, not reproducible, no subject independent evaluation, no comparison against transformer based or GNN architectures |
| | | | CNN-LSTM hybrid | | | | | | |
| Bunterngchit et al. [37], 2025 | UC San Diego rs-EEG | 31 subj (15 PD, 16 HC); 40-ch; 500 Hz; spectral power, frequency band ratios, wavelet coefficients | CTESM: CNN + Transformer blocks + LSTM layers | Train / Validation / Test split | Epoch level (not subject wise) | 99.7% | - | - | Very small dataset (31 subj), no subject independent evaluation, high accuracy likely epoch level inflated, no multi-center validation |
| GEPD: GAN + CNN [38], 2025 | UC San Diego + UNM rs-EEG | San Diego: 31 subj; UNM dataset: separate PD/HC cohort; cross-dataset generalization evaluated | GAN (data augmentation) + CNN classifier (GEPD framework) | Cross-dataset validation | Strict cross dataset subject wise | 84.3% | - | - | GAN training is unstable and computationally expensive, limited to two datasets, no transformer or SSM based temporal modeling, no medication state analysis |

**Table 3. Other Related Studies to MambaPD**

| Authors (ref), Year | Dataset Used | Dataset Statistics | Architecture / Method | Validation Strategy | Subject Split | Performance results | Limitations |
|---|---|---|---|---|---|---|---|
| Qian et al. (BioMamba) [39], 2025 | 6 public biosignal datasets (EEG, ECG) | 5 EEG + 1 ECG; binary tasks (AD, PD, workload, emotion, MI); extension adds wearable HAR | Spectro-Temporal Embedding + Bidirectional Mamba + Sparse FFN | Per-dataset (no cross-dataset); 5 seeds; 6 metrics vs 8 baselines | Subject-wise, fixed per dataset | SOTA on 5/6 datasets (all 6 metrics); TDBrain PD: 96.77% acc | TDBrain PD subset (25 PD, 25 HC, eyes-closed); single fixed split, 8 test subjects; no cross-validation; not compared with UNM/San Diego; ~53% acc on DREAMER |
| Neves et al. (EEG-ChiMamba) [40], 2026 | CAUEEG (Chung-Ang Univ. Hospital) | 1,155 subjects; HC vs MCI vs dementia; 3-class; rs-EEG | Decoupled channel wise SSM + Mamba temporal layers + Inverted LayerNorm | Strict subject-wise split; validation + held out test | Strict subject wise | 59.27% (3-class, val) | Dementia not PD, 3-class reduces accuracy vs binary PD/HC, rs-EEG paradigm transferable but not tested on PD |
| Dhavapandiammal & Palanisamy (CSFE-DAE)[41], 2026 | UC San Diego rs-EEG + independent OpenNeuro PD dataset | UCSD: 31 subjects (15 PD, 16 HC); OpenNeuro cohort size not stated in abstract | Common Spatial Pattern features + Dual-Attention EEGNet (temporal + spatial attention); Gaussian-noise augmentation | 5-fold CV + external validation on OpenNeuro | Not reported (5-fold CV) | 97.59% ± 0.47 (UCSD, 5-fold CV) | Preprint, not peer reviewed; fold split not stated as subject-wise; CNN-based, no long-range or bidirectional SSM modeling; small UCSD cohort |
| Gui et al. (EEGMamba) [42], 2024 | 8 public EEG datasets | CHB-MIT (seizure), SleepEDF (sleep), SEED (emotion), BCICIV2a (MI); variable channels/lengths | Bidirectional Mamba + ST Adaptive module + Task aware MoE | Dataset-wise per task; ablation confirms BiMamba > unidirectional | Subject wise per task | SOTA on majority of 8 datasets; linear O(n) | Never applied to PD rs-EEG, MoE adds overhead, no resting state or PD specific evaluation |
| Tegon et al. (FEMBA) [43], 2025 | TUAB + TUAR (Temple Univ. Hospital) | 21,000+ hrs unlabeled pretraining; TUAB: normal/abnormal EEG; TUAR: artifact rejection | Bidirectional Mamba SSM + self-supervised random masking; 4 model sizes (7.8M–large) | Fine-tune + held-out test per task | Subject wise held out | 81.82% bal. acc (TUAB); AUROC 0.8921 | Not evaluated on PD; requires large pretraining corpus; domain shift to PD rs-EEG unaddressed |
| Saeed et al. (EEG-SSM) [44], 2026 | OpenNeuro ds006036 (open-eyes, photic-stimulation EEG) | 88 subjects: 36 AD, 23 FTD, 29 HC; 3-class diagnosis + MMSE/duration regression | ICA preprocessing + AutoML-driven few-shot learning + SSM dual-task head (classification + severity regression) + SHAP explainability | Leave-one-subject-out (LOSO) | Subject wise (LOSO) | AD class: sens 85.9%, spec 93.8%, F1 85.5, AUC 0.98 | Dementia (AD/FTD), not PD; small cohort (88); photic-stimulation EEG, not resting state; no bidirectional selective-SSM design reported; code not yet released |
| Guo et al. (MI-Mamba) [45], 2025 | BCI Competition IV 2a + 2b | 9 subjects each; 4-class and 2-class MI; multi-channel EEG | CNN (local spatial features) + Mamba (global temporal features) + FC classifier | BCI competition protocol | Subject wise | 80.59% (2a, 4-class); 84.42% (2b, 2-class) | Motor imagery only, unidirectional Mamba, no frequency band injection, ~6x fewer params but not adapted for PD |
| Behrouz & Hashemi (Brain-Mamba) [46], 2024 | 7 real-world datasets; 3 modalities | EEG + fMRI; neurological disease detection tasks; cross-modality and cross-resolution evaluation | MLP variate fusion + S4 SSM per time-series; multi-modal neurological disease representation | Cross-dataset per modality | Subject wise per dataset | SOTA on 7 datasets across 3 modalities | Not PD specific, EEG only PD resting state not evaluated, S4 (not bidirectional Mamba) as backbone |

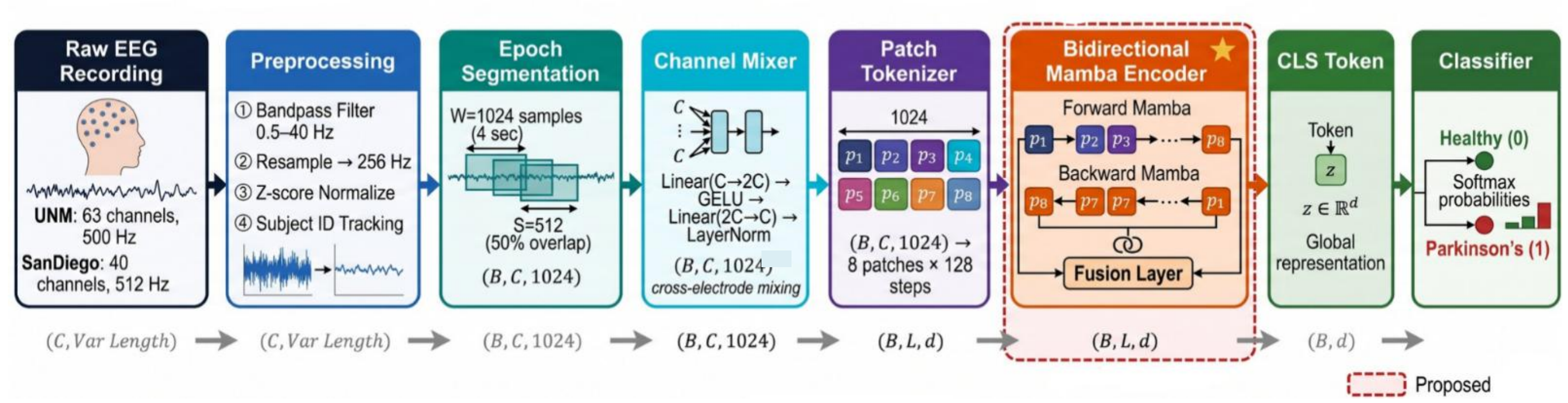

Fig. 1. Overview of the proposed MambaPD framework for resting-state EEG-based Parkinson's Disease detection.

## 3. METHODOLOGY

Figure 1 shows the end-to-end pipeline in the proposed MambaPD framework: raw EEG acquisition, raw EEG preprocessing, EEG epoch segmentation, channel mixing, patch tokenization, bidirectional Mamba encoding and binary classification. The methodology has five parts: dataset description, preprocessing, baseline architectures, the proposed MambaPD architecture, the training and evaluation.

### 3.1 Datasets

The University of New Mexico (UNM) resting-state EEG dataset was recorded at the University of New Mexico, and is among the largest publicly available Parkinson disease EEG datasets. This consists of 56 subjects: 28 clinically diagnosed Parkinson's Disease patients and 28 age matched healthy controls. EEG data were recorded with a 64-channel Brain Vision amplifier system (63 scalp EEG channels plus VEOG) with 3 auxiliary channels for the accelerometers giving 67 channels recorded at 500 Hz at the international 10-20 placement. After 12 hours of withdrawal from the dopaminergic drugs (OFF-medication condition), each PD subject participated in two separate recording sessions, one while taking the prescribed dopaminergic drug therapy (ON-medication condition) and one after the 12-hour period of medication withdrawal (OFF-medication condition). All healthy controls had a single session of rest-state recordings. In all sessions, the people sat quietly in a comfortable chair with their eyes closed for ~3 minutes to reduce motor artefacts and the neural activity associated with cognitive tasks. The raw recordings consist of 67 channels: 1-63 channels are scalp EEG electrodes to capture the electrical activity of the scalp, 64 is a channel sensitive to eye-movement artifacts (vertical electrooculography or VEOG), and 65-67 are the three channels of a triaxial accelerometer mounted on the dominant hand to separately measure tremor. The four non-cerebral channels excluded from analysis leaving 63 pure EEG channels for each recording. All epochs from PD patients were labeled as y = 1 and all healthy control epochs as y = 0. All sessions of ON and OFF are combined within one PD label.

The University of California San Diego dataset [47] has been acquired at the University of California San Diego and is made public in the standardized format of neuroimaging data known as Brain Imaging Data Structure (BIDS). It consists of 31 subjects (15 clinically diagnosed PD patients and 16 healthy controls). Recordings were made with a 40-channel amplifier system (Biosemi ActiveTwo) with a sampling rate of 512 Hz. The electrode layout is based on a modified 10-20 system that has been optimized for the Biosemi ActiveTwo headset. Two recording sessions ses-on (recorded on medication) and ses-off (recorded off medication) were completed by PD patients. Healthy controls underwent one session (ses-hc). All the recordings are resting state (eyes closed), and they are of about three minutes length. The BIDS structure contains the data in a hierarchical structure.

The two datasets vary in number of electrodes (63 vs 40), recording hardware (BrainVision vs Biosemi ActiveTwo), sampling frequency (500 vs 512 Hz), number of subjects (56 vs 31), and source of institutions (University of New Mexico vs UC San Diego). Each dataset was used separately for training and testing. No cross-dataset training and testing was implemented. The two sets of data provide two within-dataset evaluations with different recording conditions.

### 3.2 EEG Preprocessing

A standardized preprocessing pipeline was applied to both datasets using MNE-Python. Figure 2 illustrates the complete pipeline.

The UNM dataset was subjected to the same procedure: the four channels (VEOG and the three accelerometer channels) not related to EEG were excluded, leaving 63 pure scalp EEG channels. The SanDiego dataset was one of the sets that did not need any channels to be removed so it contained all 40 EEG channels. A zero-phase Hamming-windowed finite impulse response (FIR) bandpass filter was

| Table 4. Dataset Configuration | | |
|---|---|---|
| Property | UNM Dataset | SanDiego Dataset |
| Institution | Univ. of New Mexico | UC San Diego |
| Total Subjects | 56 | 31 |
| PD Patients | 28 | 15 |
| Healthy Controls | 28 | 16 |
| EEG System | BrainVision 64 ch | Biosemi Activetwo 40 ch |
| EEG Channels (used) | 63 | 40 |
| Native Sampling Rate | 500Hz | 512Hz |
| After Resampling | 256Hz | 256Hz |
| Sessions per PD | 2 (ON + OFF medication) | 2 (ses-on + ses-off) |
| Sessions per HC | 1 | 1 (ses-hc) |
| Recording Condition | Eyes closed, resting | Eyes closed, resting |
| Data Format | MATLAB .mat (EEGLAB) | BDF (BIDS) |
| Epoch Length | 4 seconds (1024 samples) | 4 seconds (1024 samples) |
| Overlap | 50% (512 sample step) | 50% (512 sample step) |
| Total Epochs | 8882 | 4475 |
| Healthy Epochs | 2995 | 1513 |
| PD Epochs | 5887 | 2962 |

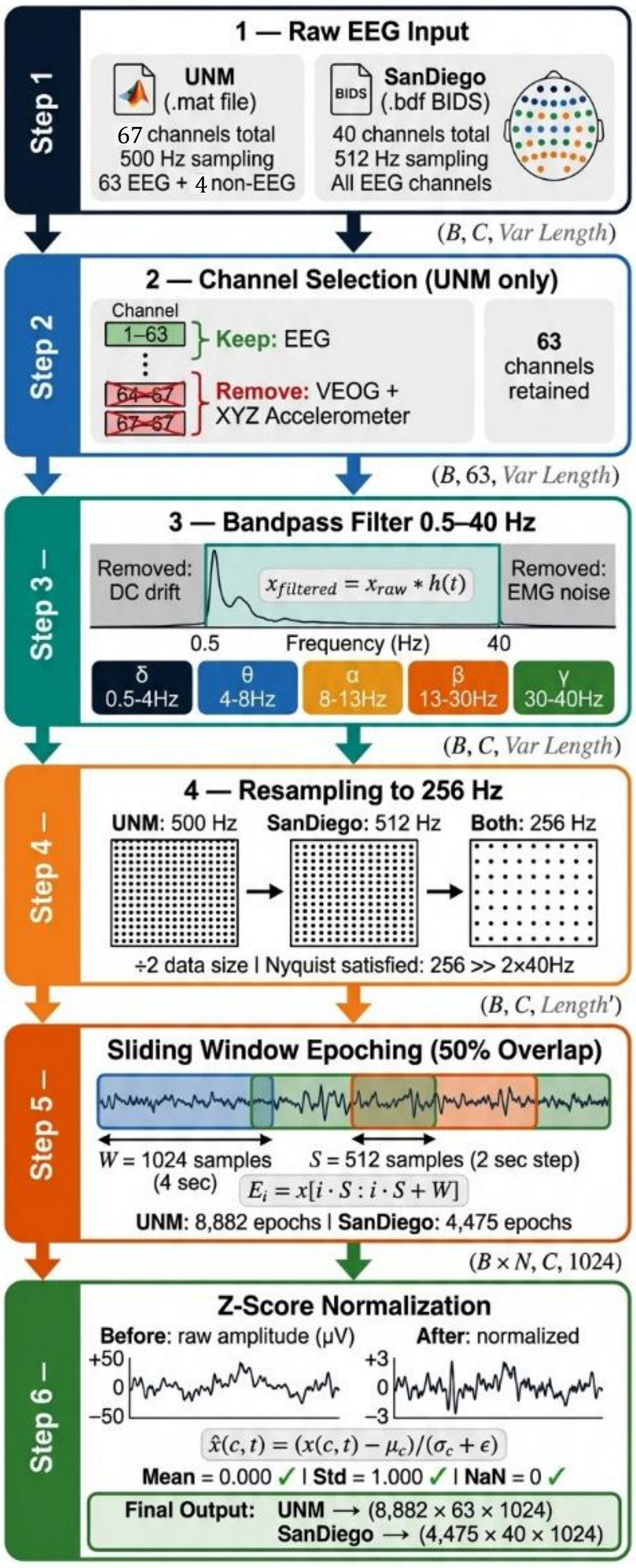

Fig. 2. EEG preprocessing pipeline applied to both datasets.

used on the data, with passband 0.5–40 Hz:

$$x_{filtered} = x_{raw}(t) * h(t) \qquad (1)$$

The Hamming-windowed FIR is denoted by h(t) and * represents linear convolution. Lower cutoff (0.5 Hz) will decrease the electrode drift and galvanic skin response. The high end of 40 Hz filters out power line (50/60 Hz) and electromyographic (EMG) interference. This passband keeps all 5 bands of frequencies that correspond to PD delta (0.5–4 Hz), theta (4–8 Hz), alpha (8–13 Hz), beta (13–30 Hz) and low gamma (30–40 Hz), Gamma above 40 Hz is not analyzed. Both sets of data were uniformly resampled to 256 Hz using MNE's FFT-based resampling. The 256 Hz sampling rate is 6.4 times the highest retained frequency (40 Hz), well above the Nyquist requirement of 80 Hz and the frequency of the signals that are diagnostically important is not aliased. Resampling will cut the dimension of the data by about half of the original rates.

The continuous preprocessed recording was divided into fixed length epochs with a sliding window:

$$E_i = x[i.S : i.S + W], \qquad i = 0,1,2, \dots \dots \qquad (2)$$

Where $x \in R^{C \times T}$ is the continuous EEG signal with C channels and T total timepoints, W=1024 samples (4 seconds at 256 Hz) is the window length, and S=512 samples (2 seconds, 50% overlap) is the step size. The window length of 4 seconds captures at least 2 full cycles of the slowest diagnostically relevant oscillation of 0.5 Hz (delta) (period 2 s). An 8,882/4,475 epoch overlap gives about double the number of training samples with no added information content at UNM and SanDiego respectively. Independently normalization per channel was done for each epoch to remove inter subject amplitude variability:

$$\hat{x}(c,t) = \frac{x(c,t) - \mu_c}{\sigma_c + \varepsilon} \quad (3)$$

Where $\mu_c = \frac{1}{W}\sum_{t=1}^{W} x(c,t)$ is the channel mean over the epoch window, $\sigma_c$ is the corresponding standard deviation, and $\varepsilon=10^{-8}$ prevents division by zero. Once normalized the mean of each channel is zero and variance is one. This normalization eliminates absolute amplitude and channel amplitude differences within each epoch (hence absolute band power), maintains the same shape of the waveforms for each channel and correlations between channels. During the pre-processing a subject identity array s was stored, in which $s_i$ contained the subject identifier for the i-th epoch. This array ensures that no epoch of any test subject is added to the training partition and at the same time that cross validation is strictly performed at patient level.

### 3.3 Baseline models

Four baseline models of deep learning sequence architectures were implemented. All receive input tensors $X \in R^{B\times C\times T}$ where B is batch size, C is EEG channels and T=1024.

**Temporal Convolutional Network (TCN)**

The stacked dilated causal convolutions with residual connections are used in the TCN. Each block expands by two dilated convolutions, such that each block is expanded (dilation factor di=2i) at the i-th block:

$$h_i = \mathrm{ELU}\left(\mathrm{BN}\left(\mathrm{Conv1D}\left(h_{i-1}, d = 2^i\right)\right)\right) + W_s h_{i-1} \quad (4)$$

where BN(·) represents batch normalization, $W_s$ represents a 1×1 convolution that matches channel dimensions and ELU(·) represents the exponential linear unit activation. The total receptive field of the stacked blocks is:

$$RF = 1 + 2(k-1)\sum_{i=0}^{n-1} 2^i = 1 + 4(2^n - 1) \quad (5)$$

=253 timestamps (k = 3, n = 6 blocks)

This is roughly 25% of the entire 1024 timestep epoch, which represents medium range temporal dependencies. The temporal dimension is reduced in global average pooling and the binary output is created in a fully connected classifier.

**Bidirectional LSTM with Temporal Attention (BiLSTM)**

The bidirectional LSTM model works with EEG sequences in both temporal directions. The forward LSTM at each time step t calculates:

$$f_t = \sigma(W_f[h_{t-1}, x_t] + b_f) \quad (6)$$

$$i_t = \sigma(W_i[h_{t-1}, x_t] + b_i) \quad (7)$$

$$\tilde{c}_t = \tanh(W_c[h_{t-1}, x_t] + b_c) \quad (8)$$

$$c_t = f_t \odot c_{t-1} + i_t \odot \tilde{c}_t \quad (9)$$

$$o_t = \sigma(W_o[h_{t-1}, x_t] + b_o) \quad (10)$$

$$h_t = o_t \odot \tanh(c_t) \quad (11)$$

where $f_t$, $i_t$, $o_t$ are forget, input and output gate activations; ct is the cell state; σ(·) is the sigmoid function and ⊙ is the element-wise multiplication operator. The reversed sequence is passed through the backward LSTM, which yields $\tilde{h}_t$. Concatenated bidirectional states $H_t=[h_t ; \tilde{h}_t] \in R^{2d}$ at each timestep are passed through a temporal attention mechanism:

$$e_t = w_a^T \tanh(W_a H_t + b_a) \quad (12)$$

$$\alpha_t = \frac{\exp(e_t)}{\sum_{t'=1}^{T}\exp(e_{t'})} \quad (13)$$

$$c = \sum_{t=1}^{T} \alpha_t H_t \quad (14)$$

where $\alpha_t$ are the normalized attention weights and c is the context vector passed to the classifier.

**EEGNet**

EEGNet consists of 3 in succession convolutional layers tailored specifically for EEG. Block 1 does a temporal convolution:

$$Z_1=\mathrm{BN}(\mathrm{Conv2D}(X_i, F_1, \mathrm{kernel}=(1,64)))$$

The depth wise spatial convolution in the C electrode channels is carried out by block 2:

$$Z_2=\mathrm{AvgPool}(\mathrm{ELU}(\mathrm{BN}(\mathrm{DepthwiseConv2D}(Z_1, F_1, \mathrm{kernel}=(C,1)))))$$

block 3 is a standard convolution over the feature maps with a temporal kernel:

$$Z_3=\mathrm{AvgPool}(\mathrm{ELU}(\mathrm{BN}(\mathrm{Conv2D}(Z_2,F_2, \mathrm{kernel}=(1,16)))))$$

with parameters $F_1$=8, D=2, $F_2$=16, The number of trainable parameters for EEGNet is roughly 6.7k.

**Patch Transformer**

The Patch Transformer tokenizes the EEG into P=8 non-overlapping patches of length $L_p$=128:

$$z_p=W_e \,.\, \mathrm{Flatten}(X(:, (p-1)L_p:pL_p))+b_e \quad (15)$$

A learnable class token $z_0$ is attached and positional encodings $E_{pos} \in R^{(P+1)\times d}$ are attached:

$$Z_0=(z_0; z_1; \ldots; z_P)+E_{pos} \quad (16)$$

For each Transformer encoder layer, I use multi-head self-attention (MHSA) and pre-layer normalization.

$$Z_l' = \mathrm{MHSA}(\mathrm{LN}(Z_{l-1})) + Z_{l-1} \quad (17)$$

$$Z_l = \mathrm{FFN}(\mathrm{LN}(Z_l')) + Z_l' \quad (18)$$

The MHSA with H heads computes:

$$\mathrm{MHSA}(Z) = \mathrm{Concat}(\mathrm{head}_1, \dots, \mathrm{head}_H)\, W^O \tag{19}$$

$$\mathrm{head}_h = \mathrm{Softmax}\left(\frac{Q_h K_h^T}{\sqrt{d_k}}\right) V_h \tag{20}$$

where $d_k = d/H$. The quadratic complexity of $O(P^2)$ with $P = 8$ gives 64 pairwise interactions (81 with the CLS token). The Patch Transformer is based on the same channel mixer and patch tokenization as MambaPD, but has a single linear patch embedding network instead of the two-layer MLP, $H = 4$ attention heads and 2 layers (Table 6)

### 3.4 Proposed Architecture: MambaPD

MambaPD is a bidirectional selective state space model for Parkinson's disease detection. It has an architecture of stages shown in figure 3

A light weight channel mixing module first extracts the cross-electrode relationships and then does temporal modelling:

$$\tilde{X} = \mathrm{LayerNorm}(W_2\, \mathrm{GELU}(W_1 X + b_1) + b_2) \tag{21}$$

These are linear layers which can operate across channels at each time point where $W_1 \in R^{2C\times C}$, $W_2 \in R^{C\times 2C}$ and $X \in R^{C\times T}$. Biases are shared across time and LayerNorm is performed across channels at each time step. The layers thus perform independent mixing of channels at each time step and do not perform mixing over time. This module allows the model to learn the PD discriminative spatial information before the temporal processing of the sequence.

The channel-mixed representation $\tilde{X} \in R^{C\times T}$ is divided into P mutually disjoint temporal patches:

$$p_i = \tilde{X}[:,\ (i-1)L_p : iL_p] \in R^{C\times L_p}, \quad i = 1, \dots, P \tag{22}$$

where $L_p$=128 timesteps (0.5 seconds) and $P=T/L_p$=8. Half a second of multi-channel EEG activity captured for each patch, corresponding to the temporal duration of fast oscillatory events of interest for PD (alpha bursts, beta spindles) but short enough so that an EEG segment is locally stationary.

Each flattened patch flattens $(p_i) \in R^{C\cdot L_p}$ is projected to embedding dimension d by a two-layer MLP:

$$e_i = \mathrm{LayerNorm}(W_4\, \mathrm{Dropout}(\mathrm{GELU}(W_3\, \mathrm{flatten}(p_i) + b_3)) + b_4) \tag{23}$$

where $W_3 \in R^{2d\times CLp}$ and $W_4 \in R^{d\times 2d}$. Learnable class token $e_0 \in R^d$ is added in front and learnable positional encodings $E_{pos} \in R(P+1)\times d$ is added:

$$E = (e_0; e_1; \dots; e_P) + E_{pos} \tag{24}$$

Selective state space model (SSM) is the backbone of MambaPD. The continuous-time model is:

$$h'(t) = A(t)\, h(t) + B(t)\, x(t) \tag{25}$$

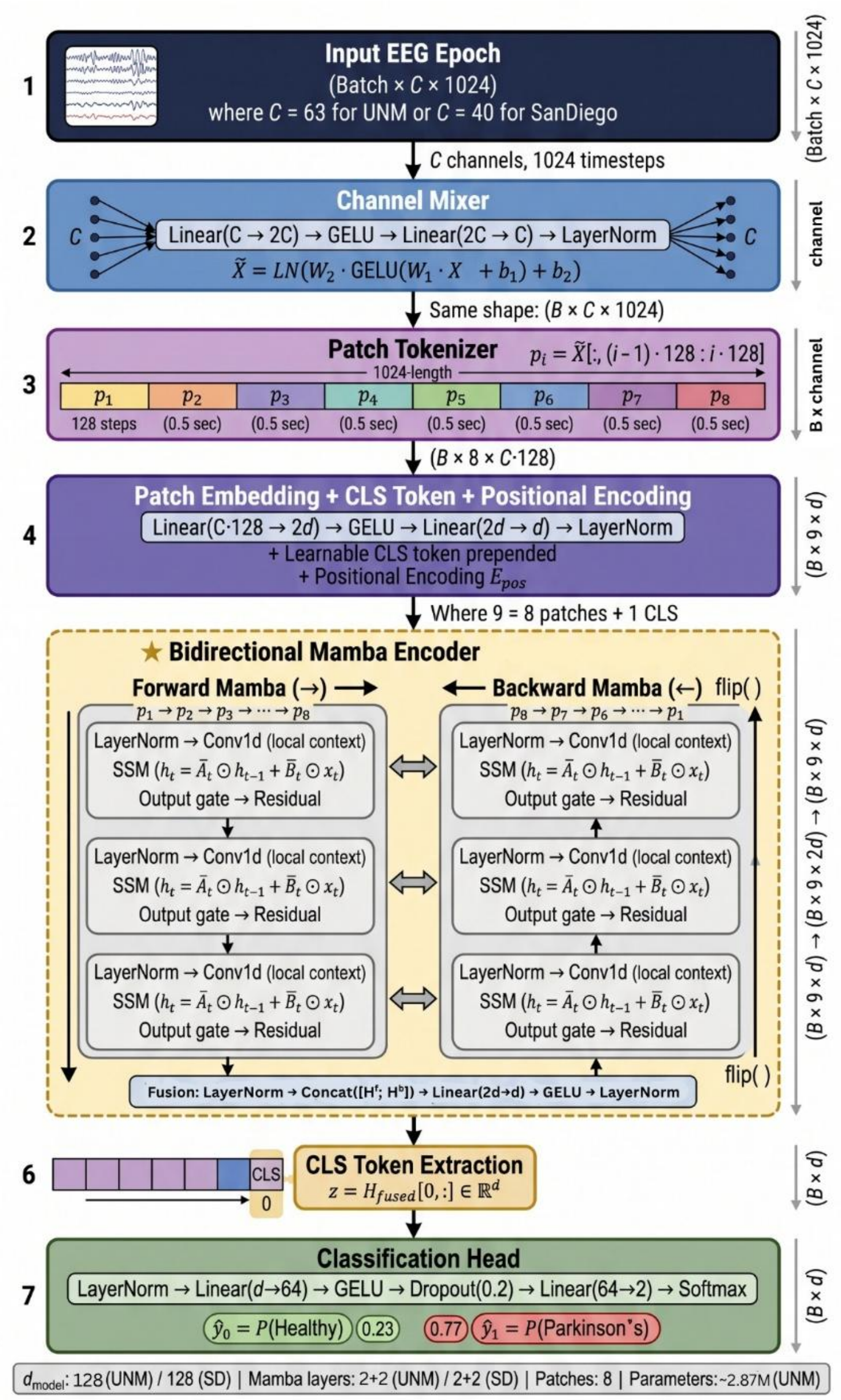

Fig. 3. Proposed MambaPD architecture for bidirectional Mamba-based Parkinson's Disease detection from resting-state EEG.

$$y(t) = C(t)\, h(t) + D\, x(t) \tag{26}$$

where $x(t) \in R^{d_{inner}}$, per channel the hidden state is $h(t) \in R^N$ (full state $h(t) \in R^{d_{inner}\times N}$), $B \in R^{N\times 1}$, $C \in R^{1\times N}$ and D is a skip connection scalar. In the selective formulation A is parameterized per channel as $A \in R^{d_{inner}\times N}$ over an expanded inner dimension $d_{inner} = E \cdot d$ (here $E = 2$).

The key innovation of Mamba over previous SSMs (e.g S4) is that B, C and the parameter Δ is a timescale which is all computed as a function of the current input, not being fixed learned constants:

$$\Delta_t = \mathrm{softplus}(\mathrm{Linear}_\Delta(x_t)) \in R^{d_{inner}} \tag{27}$$

$$B_t = \mathrm{Linear}_B(x_t) \in R^N \tag{28}$$

$$C_t = \mathrm{Linear}_C(x_t) \in R^N \tag{29}$$

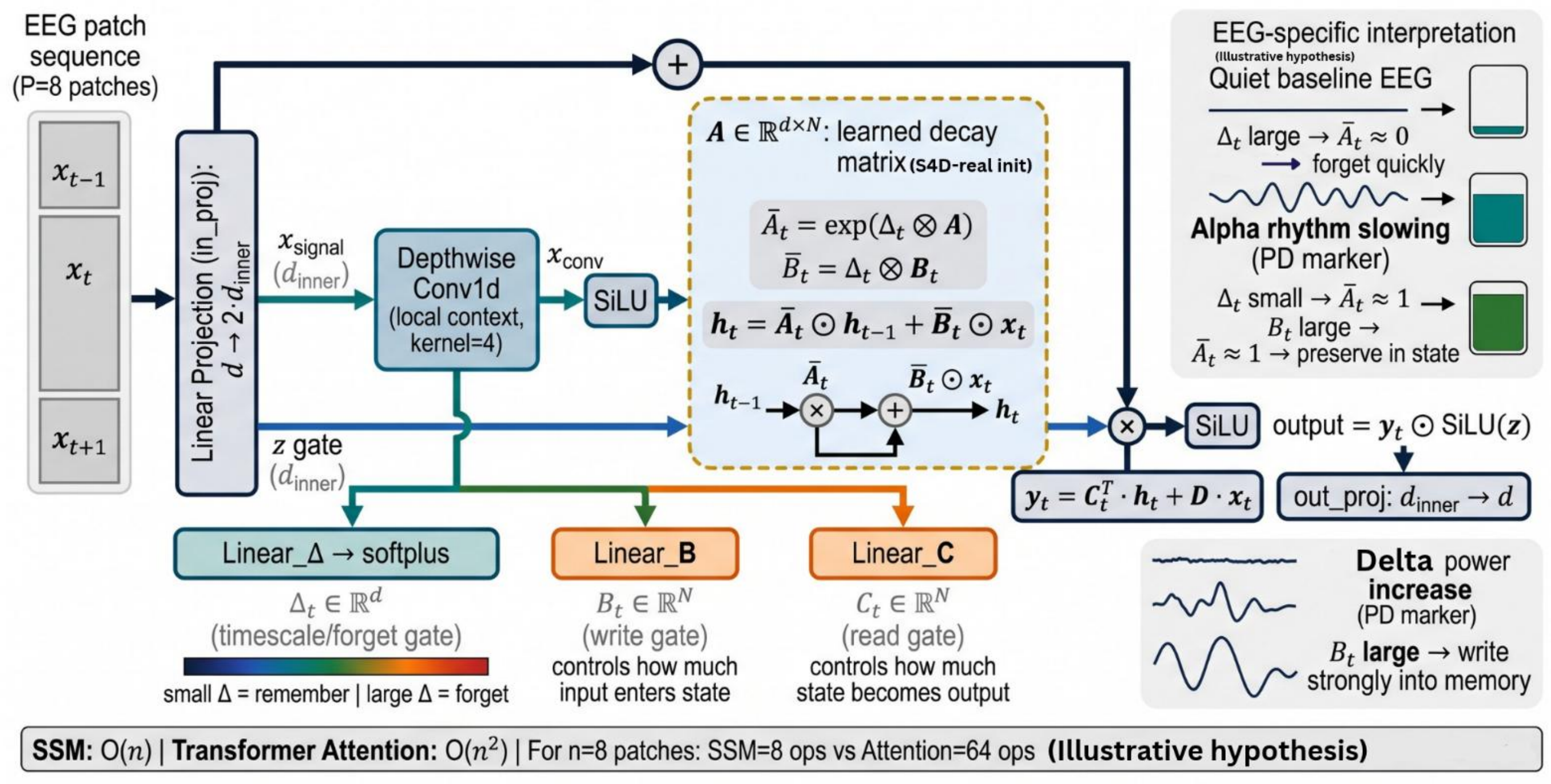

Fig. 4. *Selective state space model (SSM) mechanism in MambaPD: input-dependent gating enables adaptive EEG memory management.*

The model is input-dependent, meaning it dynamically allocates how much it can update its memory and what to read in and out of the memory, depending on the informativeness of the current EEG sample, diagnosed from it. A continuous-time system is sampled and used with a zero order hold (ZOH) method with step size Δt on the sampled EEG sequence:

$$\overline{A_t} = \exp(\Delta_t \otimes A) \quad (30)$$

$$\overline{B_t} = (\Delta_t A)^{-1}(\exp(\Delta_t A) - I)\,\Delta_t B_t \approx \Delta_t \otimes B_t \quad (31)$$

Which gives the discrete recurrence:

$$h_t = \overline{A_t} \odot h_{t-1} + \overline{B_t} \odot x_t \quad (32)$$

$$y_t = C_t^T h_t + D\, x_t \quad (33)$$

The bias of Δ projection is initialized to −2.0 and the recurrence is evaluated exactly (sequentially) over 9 tokens. The timescale parameter Δt is also used as the discretization step-size and forgetting rate. When the baseline EEG is uninformative, the hidden state should forget the past and a large Δt results in fast state decay. A small Δt retains the hidden state to be adequate in the case of a diagnostically meaningful event (alpha rhythm slowing, coherence disruption) that has taken place and whose time is to be preserved for later classification. In principle, this selective forgetting could let the model pass through uninformative patches while emphasizing spectral and coherence deviations; this is a motivation and is not tested here. Figure 4 shows the selective state space mechanism and the three input-dependent gates and their function in EEG based PD detection. Two separate stacks of Mamba blocks are used, processing the patch sequence $E \in R^{(P+1) \times d}$ in opposite time directions.

The forward Mamba stack runs patches sequentially:

$$H^f = \text{MambaStack}_{fwd}(E) \quad (34)$$

$H_i^f$ at i represents the context of patches $e_0$ to $e_i$ which is all the context that exists before patch i.The reversed sequence of patches is processed by the backward Mamba stack:

$$H^b = \text{flip}\left(\text{MambaStack}_{bwd}(\text{flip}(E))\right) \quad (35)$$

At position i, $H_i^b$ encodes context from patches $e_i$ through $e_p$ the complete future temporal context from patch i onward.

Both the direction representations are concatenated and projected in a fusion layer:

$$H_{fsd} = \text{LayerNorm}(\text{GELU}(W_f\,\text{Layernorm}([H^f; H^b]) + b^f)) \quad (36)$$

where $W_f \in R^{d \times 2d}$ and [·;·] refers to concatenation in the feature dimension. For every patch position i, the model accesses the past context (forward Mamba) and future context (backward Mamba) simultaneously.

The class token output from the fused bidirectional representation serves as the global sequence summary:

$$z = H_{fsd0} \in R^d$$

This representation is passed through the classification head

$$\hat{y} = \text{Softmax}\left(W_2\,\text{Dropout}_{p/2}\left(\text{GELU}\left(W_1 \text{Dropout}_p\ \text{LayerNorm}(z) + b_1\right)\right) + b_2\right)$$

where p=0.2, $W_1 \in R^{64 \times d}$ & $W_2 \in R^{2 \times 64}$. The predicted class is:

$$\hat{y} = \arg\max_c \hat{y}_c \in \{0\ (\text{Healthy}),\ 1\ (\text{Parkinson's Disease})\}$$

The CLS token is in position 0, thus its output in the forward stream is not providing any information about the EEG patches and, therefore, the output of the readout is effectively driven by the backward stream. With 9 tokens the linear time advantage of Mamba is not used here.

| Table 5. MambaPD configuration used for each dataset. | | |
|---|---|---|
| Component | UNM | SanDiego |
| Input shape | (B, 63, 1024) | (B, 40, 1024) |
| Channel mixer dims | 63 → 126 → 63 | 40 → 80 → 40 |
| Patch size L_p | 128 (0.5 sec) | 128 (0.5 sec) |
| Number of patches P | 8 | 8 |
| d_model | 128 | 128 |
| SSM state dim N | 16 | 16 |
| Mamba layers | 2 | 2 |
| Total Mamba layers | 4 (2 fwd + 2 bwd) | 4 (2 fwd + 2 bwd) |
| Dropout | 0.2 | 0.2 |
| Classifier hidden dim | 64 | 64 |
| Output classes | 2 | 2 |

### 3.5 Training Configuration

To solve the class imbalance, weighted cross-entropy loss with label smoothing was used:

$$\mathcal{L} = -\sum_{c\in\{0,1\}} w_c \; \tilde{y}_c \log \hat{y}_c \tag{37}$$

where $\tilde{y}_c = (1-\varepsilon)\, y_c + \frac{\varepsilon}{2}$ is the smoothed target with smoothing factor ε=0.1, and the class weight is:

$$w_c = \frac{N_{\text{total}}}{2 \times N_c} \tag{38}$$

$N_c$ is the number of epochs of class c in the dataset (computed once from all labels, not per fold). This weight gives minority healthy class (label 0) a higher weight than the majority PD class (label 1). AdamW optimizer with $\lambda=10^{-3}$, $\beta_1=0.9$ and $\beta_2=0.999$ (weight decay) were used in all models. The learning rate schedule was implemented in two phases, linear warmup for the first 10 epochs ($W_e$) between $\eta_{min}$ and $\eta_{max}$, and cosine annealing.

$$\eta(e) = \begin{cases} \eta_{min} + (\eta_{max} - \eta_{min}) \dfrac{e}{W_e} & e \le W_e \\ \eta_{min} + \dfrac{\eta_{max} - \eta_{min}}{2}\left(1 + \cos\left(\dfrac{\pi(e - W_e)}{E - W_e}\right)\right) & e > W_e \end{cases}$$

where E is the total number of training epochs. For MambaPD: $\eta_{max}=5\times10^{-4}$, $\eta_{min}=10^{-6}$, $W_e=10$, E=100 (UNM) and E=80 (SanDiego). Gradient clipping with maximum norm 1.0 was applied at every update step.

### 3.6 Evaluation Metrics

Let TP, TN, FP, FN denote true positives, true negatives, false positives, and false negatives respectively. The six evaluation metrics are:

$$\text{Accuracy} = \frac{\text{TP+TN}}{\text{TP+TN+FP+FN}} \tag{39}$$

$$\text{Precision} = \frac{\text{TP}}{\text{TP+FP}} \tag{40}$$

$$\text{Sensitivity} = \frac{\text{TP}}{\text{TP+FN}} \tag{41}$$

$$\text{Specificity} = \frac{\text{TN}}{\text{TN+FP}} \tag{42}$$

$$\text{F1}_c = \frac{2 \times \text{Precision} \times \text{Sensitivity}}{\text{Precision} + \text{Sensitivity}}$$

$$\text{F}_1 = \Sigma_c \left(\frac{\text{N}_c}{\text{N}}\right) \cdot \text{F1}_c \tag{43}$$

$$\text{AUC} = \int_0^1 \text{TPR}\left(\text{FPR}^{-1}(t)\right) dt \tag{44}$$

$$\kappa = \frac{p_o - p_e}{1 - p_e} \tag{45}$$

where $p_o$ is the observed agreement and $p_e = \sum_c p_{c,\text{pred}} \times p_{c,\text{true}}$ is the expected agreement by chance. All the metrics were calculated on the fold basis and then averaged over the 5 folds. The reported values are mean ± standard deviation from folds.

**Table 6. All models Hyperparameter configuration**

| **Hyperparameter** | **TCN** | **BiLSTM** | **EEGNet** | **Transformer** | **MambaPD** |
|---|---|---|---|---|---|
| Epochs | 100 (UNM)<br>80 (SD) | 100 (UNM)<br>80 (SD) | 100 (UNM)<br>80 (SD) | 100 (UNM)<br>80 (SD) | 100 (UNM)<br>80 (SD) |
| Batch size | 64 | 64 | 64 | 64 | 64 |
| Max LR | 1e-3 | 5e-4 | 1e-3 | 5e-4 | 5e-4 |
| Min LR | 1e-6 | 1e-6 | 1e-6 | 1e-6 | 1e-6 |
| Warmup epochs | 10 | 10 | 10 | 10 | 10 |
| Weight decay | 1e-3 | 1e-3 | 1e-3 | 1e-3 | 1e-3 |
| Dropout | 0.2 | 0.3 | 0.5 | 0.3 | 0.2 |
| Hidden dim | 64 ch | 128 | F1=8, D=2 | d=128 | 128 |
| Depth | 6 blks | 2 lyrs | F2=16 | 2 lyrs | 2+2 |
| Label smoothing | 0.1 | 0.1 | 0.1 | 0.1 | 0.1 |
| Optimizer | AdamW | AdamW | AdamW | AdamW | AdamW |
| LR schedule | Warmup+Cosine | Warmup+Cosine | Warmup+Cosine | Warmup+Cosine | Warmup+Cosine |
| Mixed precision | FP16 | FP16 | FP16 | FP16 | FP16 |
| Grad clip norm | 1.0 | 1.0 | 1.0 | 1.0 | 1.0 |

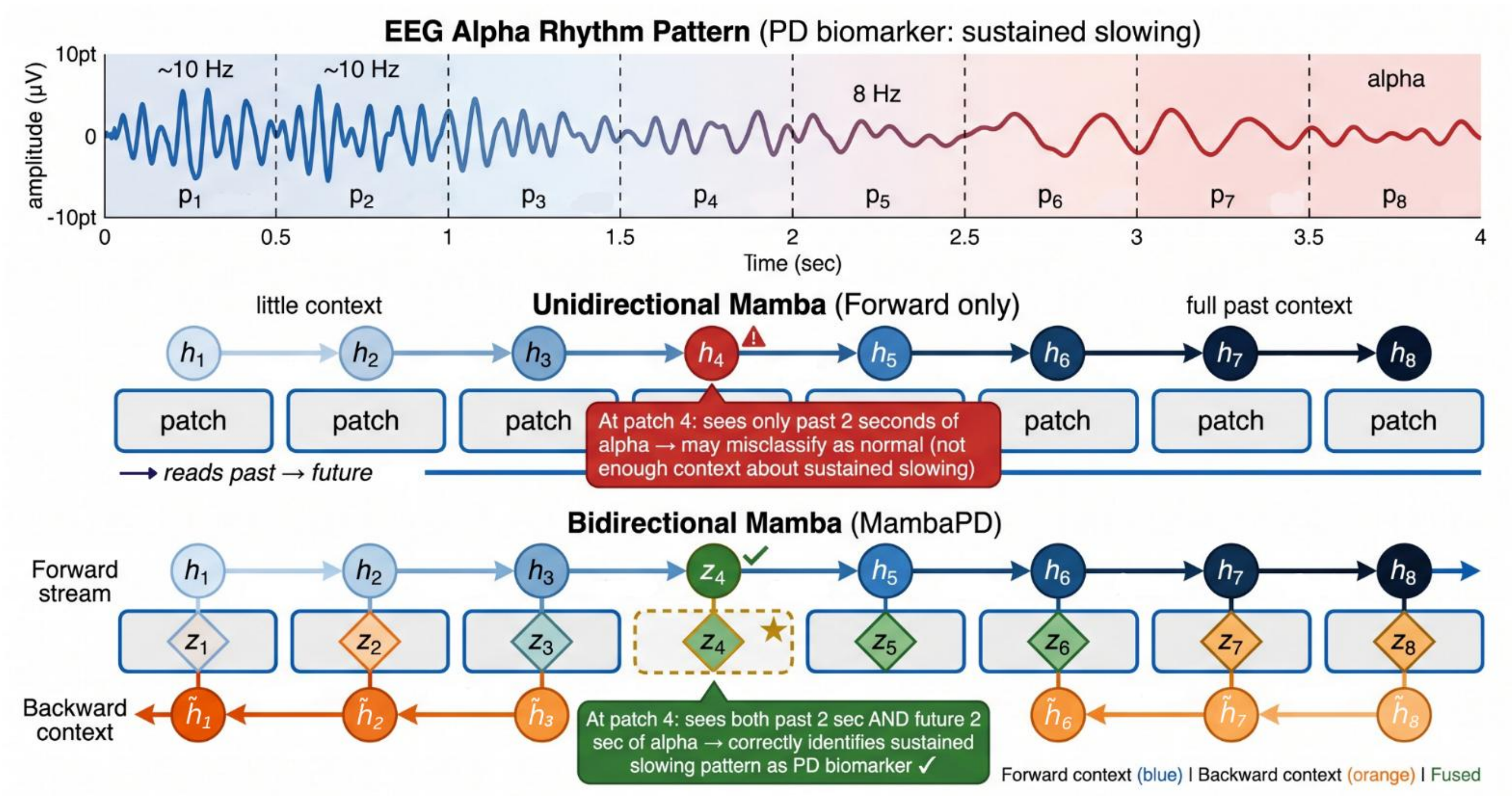

Fig. 5. Schematic illustration of bidirectional mamba processing capturing bilateral temporal context of PD relevant EEG biomarkers

5-fold stratified cross validation at the subject level is used for all experiments. Subjects not individual epochs are partitioned into five disjoint folds:

$$\mathrm{Train}_k = \{E_s : s \notin \mathrm{Fold}_k\}$$

$$\mathrm{Val}_k = \{E_s : s \in \mathrm{Fold}_k\}$$

where $E_s$ represents all epochs for the s-th subject and $\mathrm{Fold}_k$ represents all the subjects in the k-th validation fold. The PD/healthy subject ratio is approximately the same throughout folds due to stratification. Subject-level splitting does not include patient-specific EEG signatures that can easily be confused across training and testing sets, and helps to eliminate the possibility of significant over-optimistic accuracy estimates, unlike epoch-level splitting. Both sessions (ON and OFF) of a PD subject are in the same fold and the same folds were used for all models (seed 42).

# 4. EXPERIMENTAL SETUP

Data were preprocessed the same way with MNE-Python for both datasets. All PD-relevant frequency bands were preserved while all frequency bands of electrode drift and muscle noise were suppressed by means of a zero-phase Hamming-windowed FIR bandpass filter (0.5–40 Hz). Both data sets have been down-sampled to a 256 Hz sampling rate. Recordings were divided into 4 second, 50% overlapped (step size 512) epochs, resulting in 8,882 epochs from UNM and 4,475 epochs from SanDiego. The epochs were z-scored for each channel. A patient-level separation was ensured by keeping the patient identity array throughout. The training of all models was done in 16-bit mixed precision on Kaggle on an NVIDIA Tesla T4 GPU. The AdamW optimizer was applied to the problem with weight decay value $10^{-3}$ and $\beta_1 = 0.9$ and $\beta_2 = 0.999$. A linear warmup and cosine annealing learning rate schedule were used. Class imbalance was addressed using weighted cross-entropy loss with label smoothing ($\varepsilon = 0.1$), where $N_{total}$ is the total number of epochs. Gradient clipping with maximum norm 1.0 was applied. Checkpoints were selected by validation AUC, the validation set was the held-out fold, so absolute numbers are optimistic. This applies to all models.

Evaluation Protocol Strict 5-fold stratified cross validation at the subject level was used to evaluate all models. No epoch of any test patient is ever used for training at any time. For each fold six metrics were calculated and the averages computed across the folds: accuracy, weighted F1, AUC-ROC, Cohen's Kappa, sensitivity, and specificity. All models share the optimizer, schedule, weight decay, label smoothing, batch size, mixed precision and checkpoint-selection protocol. Learning rate and dropout differ by model (Table 6) and were not tuned separately.

**Table 7. Experimental results for all five models on both datasets.**

| Model | Dataset | Acc | ±std | F1 | AUC | Kappa | Sens | Spec |
|---|---|---|---|---|---|---|---|---|
| TCN | UNM | 0.7076 | 0.0969 | 0.7113 | 0.7639 | 0.3883 | 0.7439 | 0.6584 |
| TCN | SanDiego | 0.7029 | 0.0474 | 0.6957 | 0.6849 | 0.3114 | 0.8070 | 0.4918 |
| BiLSTM+Attention | UNM | 0.6991 | 0.0628 | 0.7065 | 0.7605 | 0.3673 | 0.6876 | 0.7063 |
| BiLSTM+Attention | SanDiego | 0.6669 | 0.0161 | 0.6540 | 0.7324 | 0.2529 | 0.7557 | 0.5072 |
| EEGNet | UNM | 0.6694 | 0.1032 | 0.6783 | 0.7501 | 0.3439 | 0.6276 | 0.7509 |
| EEGNet | SanDiego | 0.6007 | 0.1162 | 0.5985 | 0.6807 | 0.2136 | 0.5779 | 0.6445 |
| Transformer | UNM | 0.7146 | 0.0605 | 0.7139 | 0.7227 | 0.3492 | 0.7843 | 0.5601 |
| Transformer | SanDiego | 0.6737 | 0.0225 | 0.6656 | 0.7226 | 0.2611 | 0.7695 | 0.4941 |
| MambaPD | UNM | 0.7138 | 0.0793 | 0.7064 | 0.7647 | 0.3535 | 0.8244 | 0.5176 |
| MambaPD | SanDiego | 0.6310 | 0.0410 | 0.6305 | 0.6490 | 0.1762 | 0.7192 | 0.4611 |

# 5. RESULTS

The complementary views are presented in Figures 6–11 and the mean cross-validated metrics for all five models are listed in Table 7 on both datasets. All metrics are computed on epochs pooled within each held-out fold and then averaged over the five folds and reported as mean ± std across held out folds. The classes are imbalanced, about two thirds of the epochs are PD (Table 4). If a classifier were to classify all epochs as PD, then its accuracy would be approximately 66%, sensitivity 1.0, and specificity 0. This figure is used as a guideline for accuracy.

## 5.1 Overall Classification Performance:

UNM. The mean accuracy of each model is between 66.9% (EEGNet) and 71.5% (Patch Transformer). MambaPD reaches 71.4% (±7.9%), ahead of TCN (70.8%), BiLSTM+Attention (69.9%) and EEGNet, and 0.08 points behind the Transformer. All 5 models are 0.6-5 points higher than the always-PD reference. MambaPD achieves the best AUC (0.765) (by 0.001), and the highest sensitivity (0.824) among all methods. Its F1 (0.706) and Kappa (0.354) are lower than TCN's (0.711 and 0.388), and its specificity (0.518) is the lowest of the five models.

SanDiego. Mean accuracy ranges from 60.1% (EEGNet) to 70.3% (TCN). The best accuracy, F1 and Kappa scores are seen in the TCN model, followed by the Transformer model (67.4%) and the BiLSTM+Attention model (66.7%). MambaPD reaches 63.1% (±4.1%), ahead of EEGNet only. It has the lowest AUC (0.649), Kappa (0.176) and specificity (0.461), and its sensitivity (0.719) is fourth of five. The score for TCN increases by approximately 4 relative to the always-PD standard, while the Transformer and BiLSTM increase by around 1 each. MambaPD and EEGNet are lower than the always-PD standard. All models have weak agreement with Kappa values between 0.18 and 0.31 (with chance level at −0.02 for MambaPD's fold 2).

## 5.2 Statistical comparison

Table 7a provides paired comparisons of the accuracy per fold obtained from the MambaPD compared to the baseline networks (n = 5 folds, no multiple-comparison correction, paired t-test).

**Table 7a. Comparisons of the accuracy per fold**

| Comparison | UNM Δacc | UNM wins | UNM p | SD Δacc | SD wins | SD p |
|---|---|---|---|---|---|---|
| MambaPD − TCN | +0.006 | 3/5 | 0.85 | −0.072 | 1/5 | 0.11 |
| MambaPD − BiLSTM | +0.015 | 2/5 | 0.63 | −0.036 | 1/5 | 0.10 |
| MambaPD − EEGNet | +0.044 | 3/5 | 0.18 | +0.030 | 3/5 | 0.69 |
| MambaPD − Transformer | −0.001 | 3/5 | 0.98 | −0.043 | 1/5 | 0.09 |

No comparison reaches $p < 0.05$. The standard deviation of the paired differences between MambaPD and the Transformer on UNM is 5.6 points, which means that for 5 folds, a difference of approximately 7 points would be required to obtain a significance level $p < 0.05$. The test is not particularly sensitive to the differences here (1–5 points) so that the lack of significance does not prove that the models are not equivalent. Note that it does not imply a ranking between them. On SanDiego, MambaPD's performance is inferior to TCN, BiLSTM+Attention and Transformer for four out of five folds (p = 0.09–0.11), a trend which does not achieve statistical significance on five folds.

## 5.3 Multi-metric comparison

All six metrics are displayed as both a radar chart and a connected line chart in Figure 7 and 9. On UNM, the five models are close on accuracy (0.67–0.71), F1 (0.68–0.71), AUC (0.72–0.76) and Kappa (0.34–0.39).

**Figure 6: Accuracy and AUC-ROC Comparison**

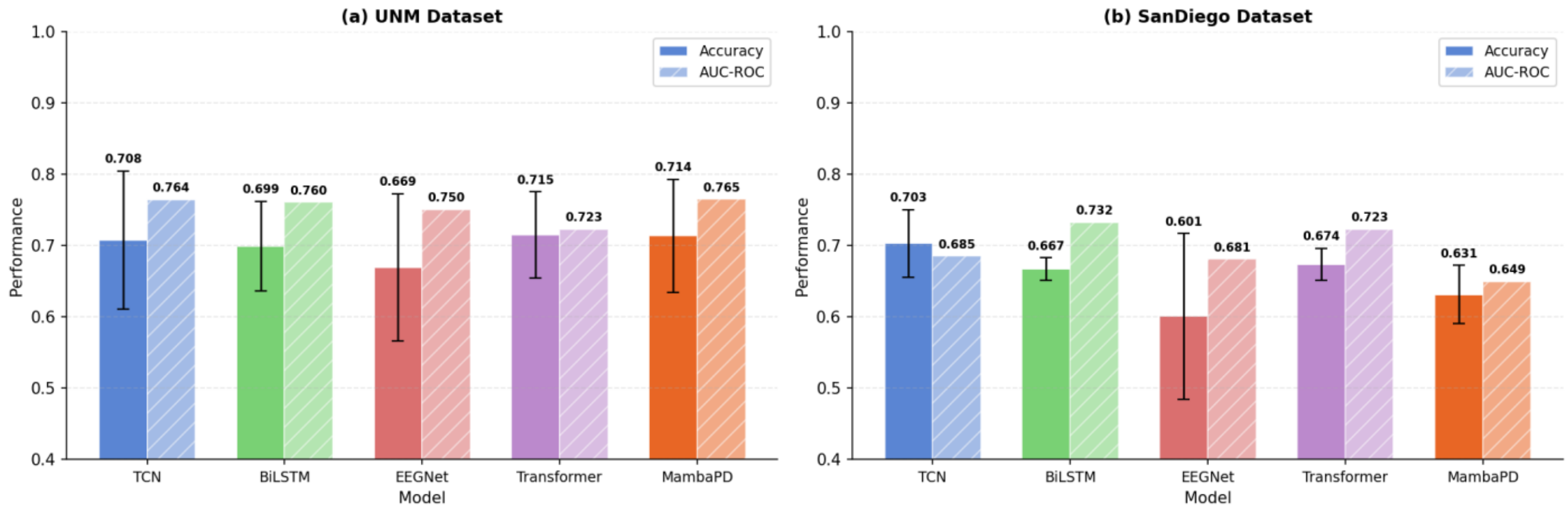

**Figure 7: Radar Chart — Multi-metric Comparison**

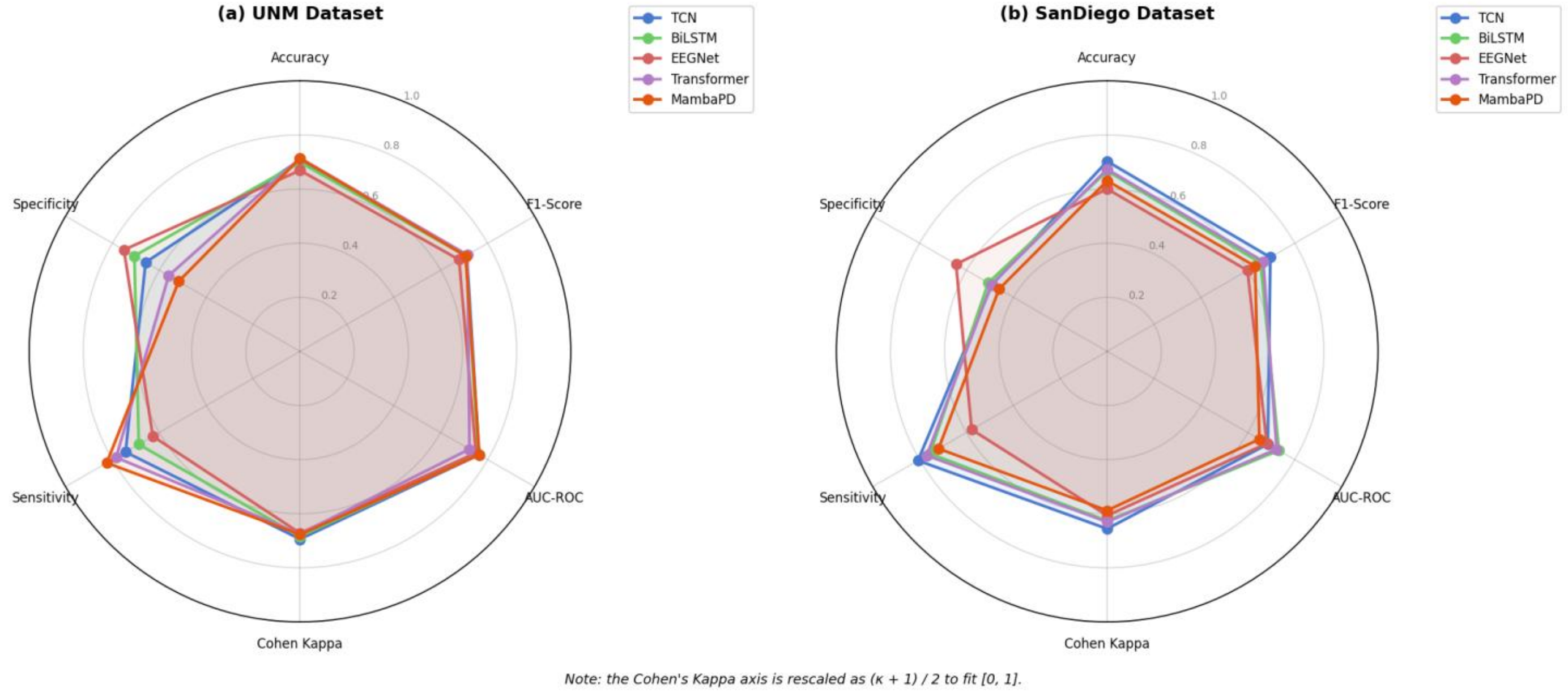

**Figure 8: Cross-validation Accuracy Distribution**

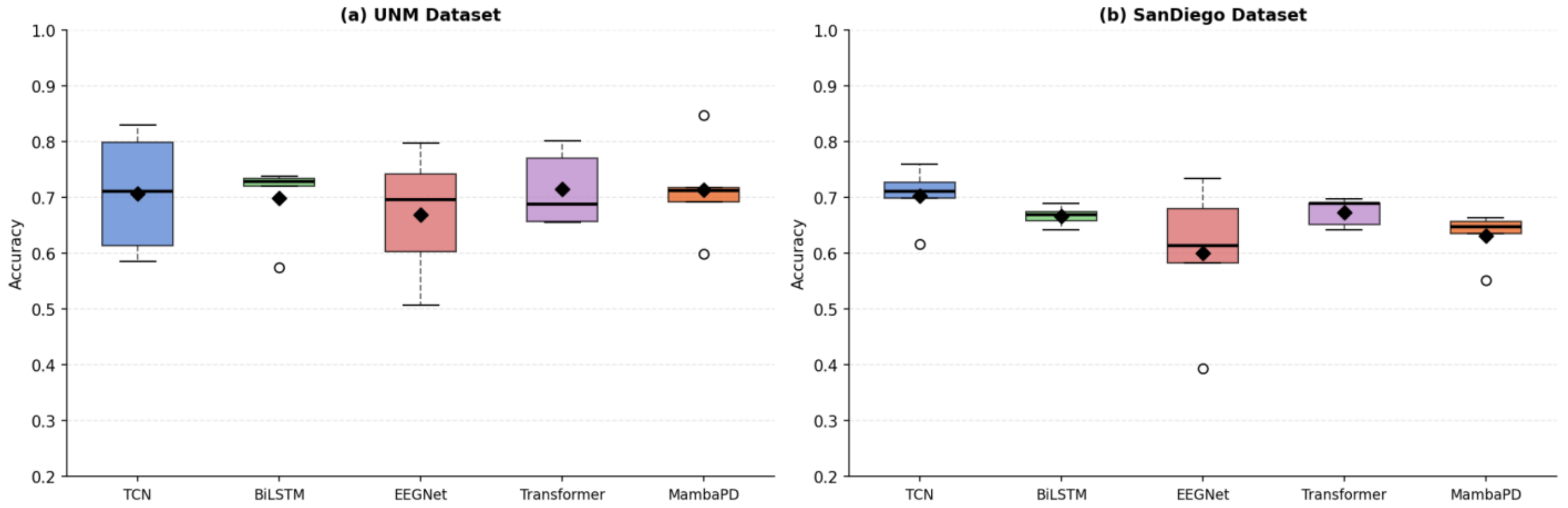

**Figure 9: All Metrics Across Models**

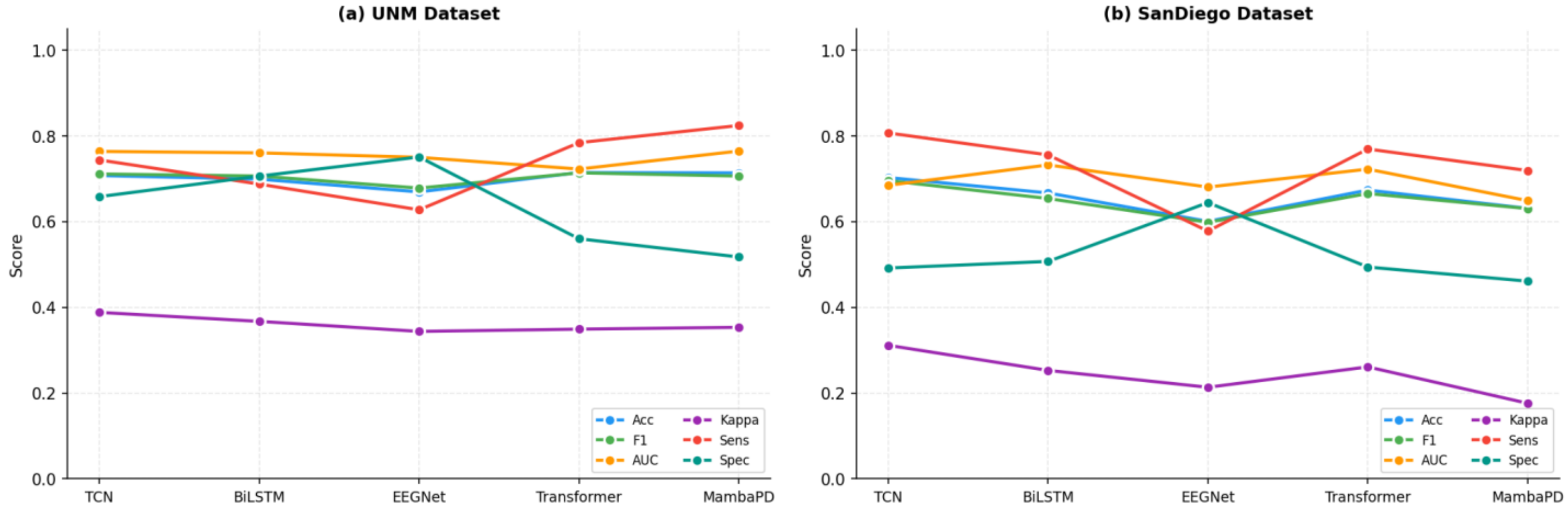

The main differences are in the sensitivity-specificity balance: MambaPD (0.824/0.518) and the Transformer (0.784/0.560) favour sensitivity, EEGNet (0.628/0.751) and BiLSTM+Attention (0.688/0.706) favour specificity, and TCN (0.744/0.658) is the most balanced. All polygons are shrunk on SanDiego. TCN is highest in accuracy, F1 and Kappa, and EEGNet is highest in specificity while MambaPD is smallest on AUC, Kappa and specificity, and fourth on sensitivity. The metrics thus concur that SanDiego is the more challenging set, and they disagree on which model is best on it, based on which metric is emphasized.

### 5.4 Fold-level behavior and stability

The accuracy per fold is displayed in Figures 8 and 11. On UNM, fold accuracy varies widely for every model (MambaPD 0.599–0.847, TCN 0.585–0.829, EEGNet 0.507–0.798, BiLSTM+Attention 0.574–0.738, Transformer 0.655–0.802).

**Figure 10: Sensitivity–Specificity Diverging Comparison**

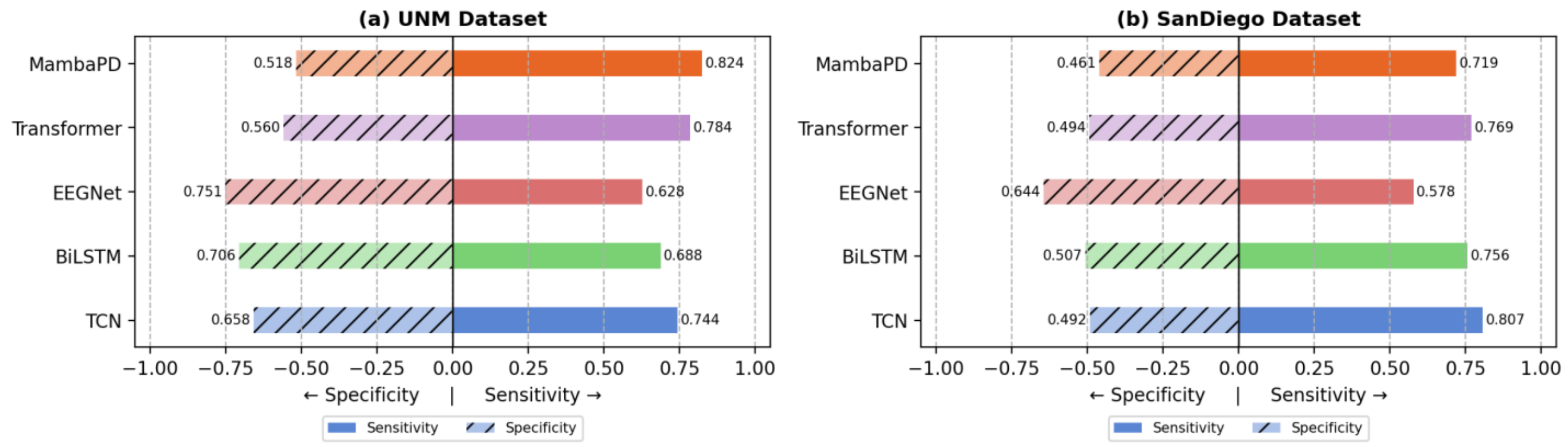

**Figure 11: Fold Accuracy Distribution (Violin + Strip)**

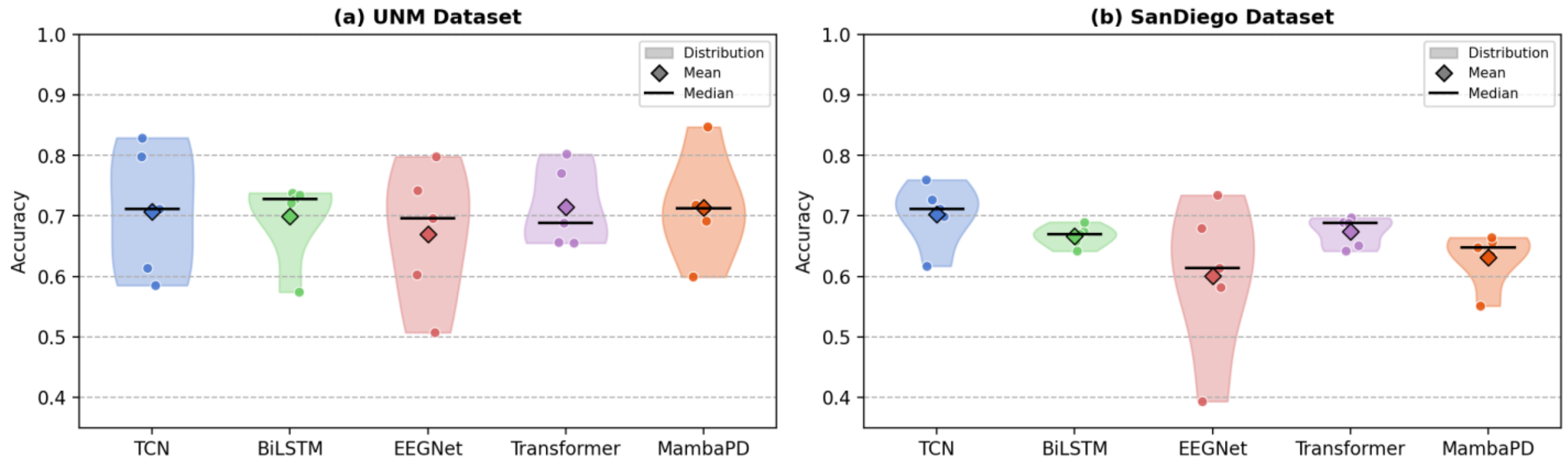

The five model means are no more than 4.5 points apart while the fold means (averaged over the five models) are as much as 21.6 points apart, with averaged fold accuracy of 0.585 (fold 1) to 0.801 (fold 3). Per-fold accuracy of different models are highly correlated (mean pairwise Pearson r = 0.76, indicative with five folds) and fold 1 is the hardest for four of five models and fold 3 the easiest for four of five. This means that the variation in the test subjects is the source of a greater amount of the variance than the model type. The fold means on SanDiego are more uniform (0.624–0.690) and there exists no consistent trend across the models (mean pairwise r ≈ 0.03). BiLSTM+Attention (std 0.016) and the Transformer (0.023) are stable, MambaPD (0.041) and TCN (0.047) are moderately variable, and EEGNet (0.116) is unstable dropping to 0.393 on one fold. The smallest fold of MambaPD is 0.551. The standard deviations are based on 6-7 test subjects per fold, meaning that the accuracy of any of the folds could move significantly when a one test subject is added or removed. Per-fold accuracies are listed in Table 8

### 5.5 Sensitivity-Specificity Trade-off

Sensitivity and specificity are displayed side by side in Figure 10; for deployment, sensitivity is an indicator of missed PD cases while specificity is an indicator of false alarms in healthy individuals. MambaPD has the lowest specificity (0.518) and the highest sensitivity (0.824) on UNM. It has the fourth highest sensitivity (0.719) on san diego out of the five (behind TCN 0.807, the Transformer 0.770, and BiLSTM 0.756) and the lowest specificity of the five. This profile might suit a screening setting where missed cases are costly but is not appropriate for confirmatory testing on UNM and is not sensitive or specific enough on SanDiego.

### 5.6 Differences between the two datasets

All five models have lower Kappa on SanDiego than on UNM and four of five have lower AUC (the Transformer is unchanged at 0.723). There are differences between the two datasets in terms of cohort size, channel count used, recording devices and number of test subjects per fold and these differences co-vary, thus I cannot argue that this difference is due to any single factor. Note that no model was trained using one dataset and tested using the other: thus the results are not a measure of cross-site generalization. The data-set also influences the ranking (TCN is highest for SanDiego but not for UNM). The biggest reduction in AUC (0.116) and Kappa (0.177) is observed in MambaPD.

### 5.7 Limitations

Absolute estimates are optimistic as AUC selected the checkpoints on the held-out fold for all models. The reported results were obtained once for each fold of each model; a second run of MambaPD on SanDiego with identical code gave 63.1% against 65.3% in the first run, so small differences between models should not be interpreted. Both datasets are small and test folds contain only 6–12 subjects. The hyperparameters were not tuned individually for each baseline; and the design choices for MambaPD were optimized on the same folds on which it was evaluated.

| Table 8. Per Fold results | | | | | | | |
|---|---|---|---|---|---|---|---|
| **Dataset** | **Model** | **Fold 1** | **Fold 2** | **Fold 3** | **Fold 4** | **Fold 5** | **Mean** |
| UNM | TCN | 0.585 | 0.613 | 0.829 | 0.799 | 0.712 | 0.708 |
| UNM | BiLSTM+Attn | 0.574 | 0.738 | 0.728 | 0.721 | 0.734 | 0.699 |
| UNM | EEGNet | 0.507 | 0.603 | 0.798 | 0.742 | 0.697 | 0.669 |
| UNM | Transformer | 0.657 | 0.655 | 0.802 | 0.771 | 0.688 | 0.715 |
| UNM | MambaPD | 0.599 | 0.718 | 0.847 | 0.713 | 0.692 | 0.714 |
| SanDiego | TCN | 0.711 | 0.727 | 0.617 | 0.760 | 0.699 | 0.703 |
| SanDiego | BiLSTM+Attn | 0.675 | 0.642 | 0.670 | 0.658 | 0.690 | 0.667 |
| SanDiego | EEGNet | 0.614 | 0.680 | 0.582 | 0.393 | 0.734 | 0.601 |
| SanDiego | Transformer | 0.690 | 0.651 | 0.697 | 0.642 | 0.689 | 0.674 |
| SanDiego | MambaPD | 0.648 | 0.551 | 0.657 | 0.664 | 0.635 | 0.631 |

Training and testing were performed separately for each dataset, there were different numbers of channels and cohorts in the datasets, and there was a different type of hardware used to record each dataset; therefore, differences from one dataset to another are not necessarily due to any single factor. The metrics are calculated on an epoch level and the decisions are taken based on a threshold of 0.5. No artifact rejection or re-referencing was applied, and per-epoch z-scoring discards absolute amplitude and band power. Medication state (ON/OFF) was not modelled. The CLS token is placed at position 0, and it has no patch data on its forward stream; thus, the patch data is effectively being driven by the backward stream. I did not make a controlled ablation and the linear-time advantage of Mamba was not used with only 9 tokens. The accuracy of approximately 60–72% is not as high as the accuracy reported on four pooled datasets for TransformEEG [26] under the nested leave-subjects-out validation (median 78.45%; 80.1% with augmentation and threshold correction), and there are differences in protocols.

## CONCLUSION

I introduced the bidirectional selective state space model for resting-state EEG PD detection (MambaPD) and compared to four baselines on two public datasets, with strict cross-validation, subject independent. MambaPD and the patch Transformer and TCN are similar on UNM (around 71%), with differences to be seen in sensitivity and specificity. On SanDiego, TCN performs best, and MambaPD is below TCN, the Transformer and BiLSTM, with the lowest specificity. The absolute accuracy is much lower than that reported with epoch level splitting and the spread across folds is larger than the spread across models on UNM. I report the pipeline and per-fold results to allow for a more stringent comparison, and I suggest multi-site validation and repeated runs, as well as nested model selection and threshold calibration in future work.